\documentclass[journal]{IEEEtran}
\usepackage{amsmath,amsfonts}
\usepackage{algorithmic}
\usepackage{algorithm}
\usepackage{array}
\usepackage[caption=false,font=normalsize,labelfont=sf,textfont=sf]{subfig}
\usepackage{textcomp}
\usepackage{stfloats}
\usepackage{url}
\usepackage{verbatim}
\usepackage{graphicx}
\usepackage{cite}

\usepackage{multirow}
\usepackage{amssymb}
\usepackage[pdftex,
            bookmarks=true,
            bookmarksnumbered=true,
            pdfstartview=FitH,
            colorlinks=true,
            linkcolor=blue,
            citecolor=blue,
            urlcolor=blue,
            pdfborder={0 0 0}
            ]{hyperref}

\begin{document}

\title{MGPC: Multimodal Network for Generalizable Point Cloud Completion With Modality Dropout and Progressive Decoding}

\author{Jiangyuan Liu, Yuhao Zhao, Hongxuan Ma, Zhe Liu, Jian Wang, and Wei Zou
\thanks{This paper was supported by the National Key Research and Development Program of China under Grant 2024YFB4708100. \textit{(Corresponding authors: Hongxuan Ma and Zhe Liu.)}}
\thanks{Jiangyuan Liu and Wei Zou are with the School of Artificial Intelligence, University of Chinese Academy of Sciences, Beijing 100043, China, and also with the State Key Laboratory of Multimodal Artificial Intelligence Systems, Institute of Automation of Chinese Academy of Sciences, Beijing 100080, China (e-mail: liujiangyuan2022@ia.ac.cn; wei.zou@ia.ac.cn).}
\thanks{Yuhao Zhao is with the State Key Laboratory of Multimodal Artificial Intelligence Systems, Institute of Automation of Chinese Academy of Sciences, Beijing 100080, China, and with the Chemical Defense Institute, Academy of Military Sciences, Beijing 102205, China (e-mail: zhaoyuhao2021@ia.ac.cn).}
\thanks{Hongxuan Ma is with the State Key Laboratory of Multimodal Artificial Intelligence Systems, Institute of Automation of Chinese Academy of Sciences, Beijing 100080, China (e-mail: hongxuan.ma@ia.ac.cn).}
\thanks{Zhe Liu is with the Chemical Defense Institute, Academy of Military Sciences, Beijing 102205, China (e-mail: zheliu\_academia@hotmail.com).}
\thanks{Jian Wang is with the Key Laboratory of Cognition and Decision Intelligence for Complex Systems, Institute of Automation of Chinese Academy of Sciences, Beijing 100080, China (e-mail: jianwang@ia.ac.cn).}
\thanks{The dataset and model are available at \url{https://github.com/L-J-Yuan/MGPC}.}
}



\maketitle

\begin{abstract}
Point cloud completion aims to recover complete 3D geometry from partial observations caused by limited viewpoints and occlusions. Existing learning-based works, including 3D Convolutional Neural Network (CNN)-based, point-based, and Transformer-based methods, have achieved strong performance on synthetic benchmarks. However, due to the limitations of modality, scalability, and generative capacity, their generalization to novel objects and real-world scenarios remains challenging. In this paper, we propose MGPC, a generalizable multimodal point cloud completion framework that integrates point clouds, RGB images, and text within a unified architecture. MGPC introduces an innovative modality dropout strategy, a Transformer-based fusion module, and a novel progressive generator to improve robustness, scalability, and geometric modeling capability. We further develop an automatic data generation pipeline and construct MGPC-1M, a large-scale benchmark with over 1,000 categories and one million training pairs. Extensive experiments on MGPC-1M and in-the-wild data demonstrate that the proposed method consistently outperforms prior baselines and exhibits strong generalization under real-world conditions.
\end{abstract}


\section{Introduction}
\IEEEPARstart{A}{s} a widely used 3D representation \cite{point2014}, point clouds play a crucial role in spatial perception and measurement systems. Despite their popularity, point clouds captured by modern 3D sensors, such as depth cameras and LIDAR scanners, are often incomplete and sparse due to limited viewpoints, occlusions, and inherent sensing uncertainties \cite{survey22022}. Recovering complete object geometry from partial observations is therefore a fundamental problem and is critical for many downstream applications, including pose estimation \cite{gernet2023}, manipulation \cite{giga2021}, and industrial inspection \cite{oil2024}.

\begin{figure}[!t]
\centering
\includegraphics[width=0.49\textwidth]{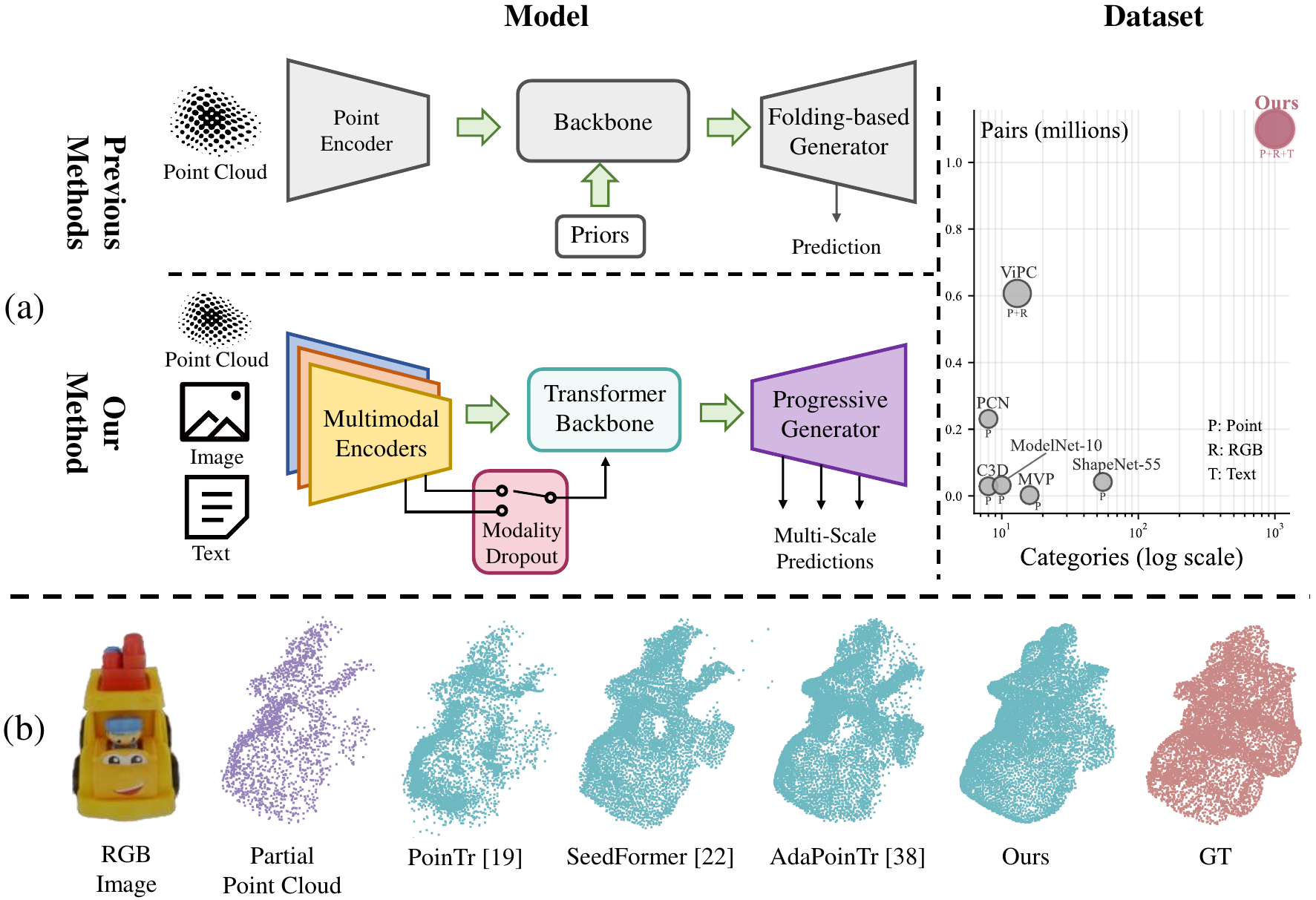}
\caption{(a) Overview of the proposed framework and dataset. Our method encodes point clouds, RGB images, and text using modality-specific encoders, followed by a scalable Transformer with modality dropout, and predicts completed point clouds via a progressive generator with multi-scale outputs. The proposed multimodal benchmark substantially increases both category coverage and paired-sample scale compared to existing datasets. (b) Qualitative comparison on a novel real-world sample. Methods pretrained on prior datasets struggle to generalize, producing incomplete or noisy reconstructions.}
\label{fig1}
\end{figure}

With the rapid advancement of deep learning, point cloud completion has been increasingly studied in a data-driven fashion. Early approaches process voxelized representations using 3D CNNs \cite{dai2017shape, grnet2020, pvd2021}, which suffer from high computational cost and limited scalability as resolution increases. Moving beyond voxel grids, PointNet \cite{pointnet2017} and its variants \cite{pointnet++2017, dgcnn2019, epnet2024} inspired subsequent works \cite{pcn2018, crn2020, odgnet2024, msn2020, nsfa2020, pmp2021} to directly operate on unordered point sets through symmetric aggregation functions such as max-pooling. More recently, Transformer-based architectures \cite{pointr2021, proxyformer2023, snowflakenet2021, seedformer2022} have been introduced to better capture long-range dependencies within point clouds. Although these methods achieve strong performance on existing synthetic benchmarks, their generalization ability to novel objects and real-world scenarios remains limited \cite{survey2023, robust2022}. As shown in Fig. \ref{fig1}(b), state-of-the-art models trained on existing datasets often produce incomplete shapes and scattered outliers on a previously unseen real-world object.

Through systematic analysis, we identify three key factors that hinder the generalization of existing point cloud completion methods: (1) \textbf{Limited modality}. Most prior works \cite{pcn2018, pointr2021, seedformer2022} rely solely on partial point clouds as input. However, geometry alone may be insufficient to resolve shape ambiguity, as distinct objects can exhibit similar geometric structures under certain viewpoints. Incorporating complementary modalities, such as RGB images and textual descriptions, provides additional appearance and semantic cues that are beneficial for robust generalization. (2) \textbf{Limited scalability}. Due to the relatively small scale of existing datasets, many methods depend on strong inductive biases, including local operators (e.g., CNNs \cite{dai2017shape, grnet2020, pvd2021} and KNN-based modules \cite{pointr2021, proxyformer2023, snowflakenet2021, seedformer2022}), handcrafted shape priors \cite{ppcc2024, paco2025, tcorresnet2024, dcpcn2025}, or carefully engineered network components \cite{einet2024, pfnet2020, crapcn2024}, as summarized in the first row of Fig. \ref{fig1}(a). While effective in low-data regimes, such designs may constrain scalability and reduce the gains achievable as data scale increases \cite{openshape2023, mcc2023, lrm2023, ptv32024}. (3) \textbf{Limited generative capacity}. The generative capacity of the generator largely determines the fidelity of completed point clouds. Many existing methods adopt folding-based generators \cite{foldingnet2018, pcn2018, pointr2021} or MLP-based generators \cite{adapointr2023, pfnet2020}, which deform a 2D grid into 3D surfaces or regress coordinates in a one-shot manner. Such parameterizations often have limited expressiveness for modeling complex 3D geometric distributions, which can hinder fine-detail recovery and lead to suboptimal reconstruction quality.

To address the above challenges, we propose \textbf{MGPC}, a simple yet effective \underline{M}ultimodal framework for \underline{G}eneralizable \underline{P}oint cloud \underline{C}ompletion. As illustrated in the second row of Fig. \ref{fig1}(a), MGPC encodes point clouds, RGB images, and text into modality-specific tokens and fuses them using a scalable attention-based backbone. To reduce overfitting and improve robustness when auxiliary modalities are unavailable at inference time, a novel modality dropout strategy for multimodal conditioning is introduced. Finally, instead of conventional folding-based or MLP-based generators, MGPC adopts a progressive generator with multi-scale supervision to refine geometry in a coarse-to-fine manner and enhance reconstruction fidelity.

Moreover, to fully exploit the scalability of the Transformer architectures and further enhance generalization, we develop an automatic data generation pipeline that incorporates a large Vision-Language Model (VLM) for data filtering, along with more realistic data normalization strategies. Based on this pipeline, we construct \textbf{MGPC-1M}, the largest point cloud completion benchmark to date comprising over 1,000 categories and 1 million paired samples. Extensive experiments on this benchmark demonstrate that MGPC consistently outperforms both single-modal and cross-modal competitors by a substantial margin. Additional in-the-wild evaluations on real-world data further validate the strong zero-shot generalization capability of the proposed approach.

In summary, our main contributions are as follows:
\begin{itemize}
\item{We propose MGPC, a unified multimodal framework for point cloud completion that jointly leverages point clouds, RGB images, and text to improve generalization.}
\item{We introduce modality dropout, together with a Transformer-based fusion module and a progressive generator, to enhance robustness, scalability, and reconstruction fidelity.}
\item{We present an automatic data generation pipeline and construct MGPC-1M, a large-scale benchmark with over 1,000 categories and 1 million training samples.}
\item{The proposed method achieves state-of-the-art performance on large-scale benchmarks and real-world scenarios, demonstrating strong generalization.}
\end{itemize}

\section{RELATED WORK}
\subsection{Single-Modal Point Cloud Completion}
The majority of existing point cloud completion methods take partial point clouds as the sole input modality \cite{pcn2018, pointr2021, seedformer2022}. These works can be broadly categorized into 3D CNN-based methods \cite{dai2017shape, grnet2020, pvd2021}, point-based methods \cite{pcn2018, crn2020, odgnet2024, msn2020, nsfa2020, pmp2021}, and Transformer-based methods \cite{pointr2021, proxyformer2023, snowflakenet2021, seedformer2022}.

Early approaches \cite{dai2017shape, grnet2020, pvd2021} voxelize point clouds into regular grids and apply 3D CNNs in a manner analogous to 2D image processing. While effective at low resolutions, voxel-based representations incur prohibitive computational and memory costs as resolution increases, which severely limits their scalability. Motivated by the success of PointNet \cite{pointnet2017} and its variants \cite{pointnet++2017, dgcnn2019, epnet2024}, PCN \cite{pcn2018} and subsequent point-based methods \cite{crn2020, odgnet2024, msn2020, nsfa2020, pmp2021} operate directly on unordered point sets via shared MLPs, typically coupled with folding-based decoders \cite{foldingnet2018}. For example, CRN \cite{crn2020} introduces a two-stage pipeline that refines coarse predictions through a lifting module, while ODGNet \cite{odgnet2024} employs an orthogonal dictionary to guide training. However, both approaches require learning dataset-specific shape priors, which may restrict generalization to unseen objects. MSN \cite{msn2020} and NSFA \cite{nsfa2020} formulate point cloud generation as a deformation process, whereas TopNet \cite{topnet2019} and PMPNet \cite{pmp2021} model the generation process using tree structures and earth mover formulations, respectively. Despite their effectiveness, many of these methods depend on tightly constrained generators with limited representational capacity, which can lead to oversmoothed surfaces and suboptimal reconstruction quality.

Inspired by advances in natural language processing and 2D vision, recent studies have introduced Transformer architectures \cite{transformer2017} to capture global dependencies within point clouds. As a pioneering work, PoinTr \cite{pointr2021} proposes a geometric-aware Transformer that incorporates KNN operators to exploit local geometric priors. Subsequent works largely follow two main directions: introducing additional inductive biases \cite{ppcc2024, paco2025, tcorresnet2024, dcpcn2025} and designing specialized architectural components \cite{einet2024, pfnet2020, crapcn2024}. In the first line, PPCC \cite{ppcc2024} and PaCo \cite{paco2025} investigate completion based on object parts and planar primitives, respectively. TCorresNet \cite{tcorresnet2024} and DCPCN \cite{dcpcn2025} resort to external data structures such as spherical templates and interfaces, which often require extensive pre-processing and hyperparameter tuning. In the second line, CRA-PCN \cite{crapcn2024} and PointAttn \cite{pointattn2024} study local-global fusion strategies, while SeedFormer \cite{seedformer2022} and ProxyFormer \cite{proxyformer2023} focus on modifications of Transformer architectures tailored to missing-part sensitivity and upsampling. Despite these efforts, the increasing reliance on complex inductive biases and intricate module designs may limit scalability and generalization under large-scale data regimes \cite{ptv32024, lrm2023}. In contrast, we shift our focus to more scalable design based on pure self-attention and cross-attention, combined with large-scale training \cite{openshape2023, automatic2019} and further exploit complementary information from multiple modalities to improve completion quality.

\subsection{Cross-Modal Point Cloud Completion}
Cross-modal point cloud completion methods include multi-view approaches \cite{xiao2022, svdformer2023, aednet2024, sparsenet2021, mmdr2025} and image-assisted approaches \cite{vipc2021, xmfnet2022, csdn2023, egiinet2024, bfr2025}. Multi-view methods, such as SVDFormer \cite{svdformer2023} and AEDNet \cite{aednet2024}, follow a common paradigm of projecting partial point clouds into depth images from multiple viewpoints, which are then used as auxiliary inputs. SparseNet \cite{sparsenet2021} and MMDR \cite{mmdr2025} further introduce rendering-based losses by comparing projections of the completed point clouds with Ground-Truth (GT) projections. However, in these approaches, the image modality is derived directly from the input point cloud itself, and thus does not provide additional external information beyond the original geometry.

Another line of research incorporates RGB images to assist point cloud completion \cite{vipc2021, xmfnet2022, csdn2023, egiinet2024, bfr2025}. ViPC \cite{vipc2021} addresses cross-modal learning by training a generative network to regress complete point clouds from single-view images. Nevertheless, bridging RGB apperance to precise 3D geometry is non-trivial, which may limit ViPC in recovering accurate and complete geometry, especially for occluded regions. Subsequent methods, such as XMFNet \cite{xmfnet2022} and CSDN \cite{csdn2023}, improve cross-modal fusion in the latent space using cross-attention mechanisms and affine normalization, respectively. EGIINet \cite{egiinet2024} introduces a feature transfer loss to explicitly guide the modality alignment. More recently, BfR \cite{bfr2025} explores a retrieval-based paradigm that leverages similar reference samples from a database. However, its performance heavily depends on the quality and scale of the database and incurs additional computational overhead during inference. In contrast to existing approaches that primarily focus on sophisticated modality fusion strategies, we adopt a unified framework that integrates point clouds, RGB images, and textual descriptions for point cloud completion. Furthermore, the proposed modality dropout mechanism effectively reduces overfitting while improving robustness and flexibility under missing-modality scenarios.

\begin{table*}[!t]
\centering
\caption{Summary and Comparison of Point Cloud Completion Datasets}
\label{tab1}
\setlength{\tabcolsep}{4.5pt}
\begin{tabular}{c|c|cc|cc|c|c|c|ccc}
\hline
\multirow{2}{*}{} & \multirow{2}{*}{Category} & \multicolumn{2}{c|}{Training Set}                & \multicolumn{2}{c|}{Testing Set}   & \multirow{2}{*}{Modality}                & \multirow{2}{*}{Domain}                                       & \multirow{2}{*}{Normalization} & \multicolumn{3}{c}{Virtual Camera}                                                                                                   \\ \cline{3-6} \cline{10-12}
                  &                       & \multicolumn{1}{c|}{Model}        & Pair         & \multicolumn{1}{c|}{Model}  & Pair &       &                                                               &                        & \multicolumn{1}{c|}{Generation}         & \multicolumn{1}{c|}{Views} & Distribution        \\ \hline
KITTI \cite{kitti2013}             & 1                     & \multicolumn{1}{c|}{/}            & /            & \multicolumn{1}{c|}{2.4k}   & 2.4k & P          & Real                                                          & Bbox-Centric              & \multicolumn{1}{c|}{LiDAR}              & \multicolumn{1}{c|}{1}     & Random                                                                           \\
ModelNet-10 \cite{modelnet2015}       & 10                    & \multicolumn{1}{c|}{4k}           & 32k          & \multicolumn{1}{c|}{1k}     & 0.2k & P          & Synthetic                                                          & $\sim$                 & \multicolumn{1}{c|}{Back-Projection}          & \multicolumn{1}{c|}{8}     & Random                                                                       \\
PCN \cite{pcn2018}              & 8                     & \multicolumn{1}{c|}{29k}          & 231k         & \multicolumn{1}{c|}{1.2k}   & 1.2k & P          & Synthetic                                                          & GT-Centric                & \multicolumn{1}{c|}{Back-Projection}          & \multicolumn{1}{c|}{8}     & Random                                                                   \\
C3D \cite{topnet2019}              & 8                     & \multicolumn{1}{c|}{29k}          & 29k          & \multicolumn{1}{c|}{1.2k}   & 1.2k & P          & Synthetic                                                          & GT-Centric                & \multicolumn{1}{c|}{Back-Projection}          & \multicolumn{1}{c|}{1}     & Random                                                                         \\
MVP \cite{vrcnet2021}              & 16                    & \multicolumn{1}{c|}{2.4k}         & 62k          & \multicolumn{1}{c|}{1.6k}   & 41k  & P          & Synthetic                                                          & /                      & \multicolumn{1}{c|}{Back-Projection}          & \multicolumn{1}{c|}{26}    & Uniform                                                                         \\
ShapeNet-34 \cite{pointr2021}       & 55                    & \multicolumn{1}{c|}{46k}          & 46k          & \multicolumn{1}{c|}{5.7k}   & 5.7k & P          & Synthetic                                                          & GT-Centric                & \multicolumn{1}{c|}{Crop}               & \multicolumn{1}{c|}{1}     & Random                                                                         \\
ShapeNet-55 \cite{pointr2021}      & 55                    & \multicolumn{1}{c|}{42k}          & 42k          & \multicolumn{1}{c|}{10k}    & 10k  & P          & Synthetic                                                          & GT-Centric                & \multicolumn{1}{c|}{Crop}               & \multicolumn{1}{c|}{1}     & Random                                                                         \\
ViPC \cite{vipc2021}             & 13                    & \multicolumn{1}{c|}{38k}          & 607k         & \multicolumn{1}{c|}{$\sim$} & 152k & P \& I          & Synthetic                                                          & GT-Centric                & \multicolumn{1}{c|}{Occlusion}          & \multicolumn{1}{c|}{24}    & Uniform                                                                        \\ \hline
\textbf{Ours}     & \textbf{1k+}          & \multicolumn{1}{c|}{\textbf{53k}} & \textbf{1M+} & \multicolumn{1}{c|}{1k}     & 20k  & \multicolumn{1}{c|}{\textbf{P \& I \& T}}  & \textbf{\begin{tabular}[c]{@{}c@{}}Synthetic\\ +Real\end{tabular}} & \textbf{Input-Centric}    & \multicolumn{1}{c|}{\textbf{Back-Projection}} & \multicolumn{1}{c|}{20}    & \textbf{Uniform}  \\ \hline

\multicolumn{12}{l}{\footnotesize
\textit{Note:} P, I, and T indicate point cloud, image, and text modalities, respectively.
}

\end{tabular}
\end{table*}

\subsection{Point Cloud Completion Datasets}
Dataset plays a crucial role for learning-based point completion methods \cite{openshape2023}. Early efforts leverage 3D models from the ModelNet \cite{modelnet2015} dataset to evaluate shape completion. PCN \cite{pcn2018} subsequently introduces the first widely adopted benchmark based on high-quality ShapeNet \cite{shapenet2015} models. Later datasets expand upon PCN by incorporating more categories and viewpoints. In particular, ShapeNet-34 and ShapeNet-55 \cite{pointr2021} utilize all 55 categories from ShapeNetCore \cite{shapenet2015} and stratify test samples according to different difficulty levels. However, the crop-based strategy used to generate partial point clouds deviates from realistic sensing processes of depth cameras and laser scanners. The MVP dataset \cite{vrcnet2021} addresses this limitation by rendering partial observations using diverse and uniformly distributed camera poses, although the number of paired samples remains relatively limited. KITTI \cite{kitti2013} is frequently employed to evaluate real-world generalization, but it contains only vehicle point clouds and lacks ground-truth complete shapes. More recently, ViPC \cite{vipc2021} constructs a large-scale dataset with paired RGB images to facilitate cross-modal learning.

Overall, existing point cloud completion datasets are limited in terms of category diversity and paired data scale \cite{survey22022}, as summarized in Table \ref{tab1}. Most datasets rely on synthetic ShapeNet \cite{shapenet2015} models and adopt suboptimal configurations, such as ground-truth-centric normalization and crop-based partial generation. These limitations contribute to the poor real-world generalization observed in current methods \cite{survey2023}. To address these issues, we construct a new large-scale dataset based on Objaverse \cite{objaverse2023} and GSO \cite{gso2022}, featuring significantly increased scale and more realistic data generation settings. Our dataset comprises over 1,000 categories and 1 million paired samples, and we expect it to serve as a strong foundation for improving generalization and advancing research in point cloud completion.

\begin{figure*}[!t]
\centering
\includegraphics[width=0.98\textwidth]{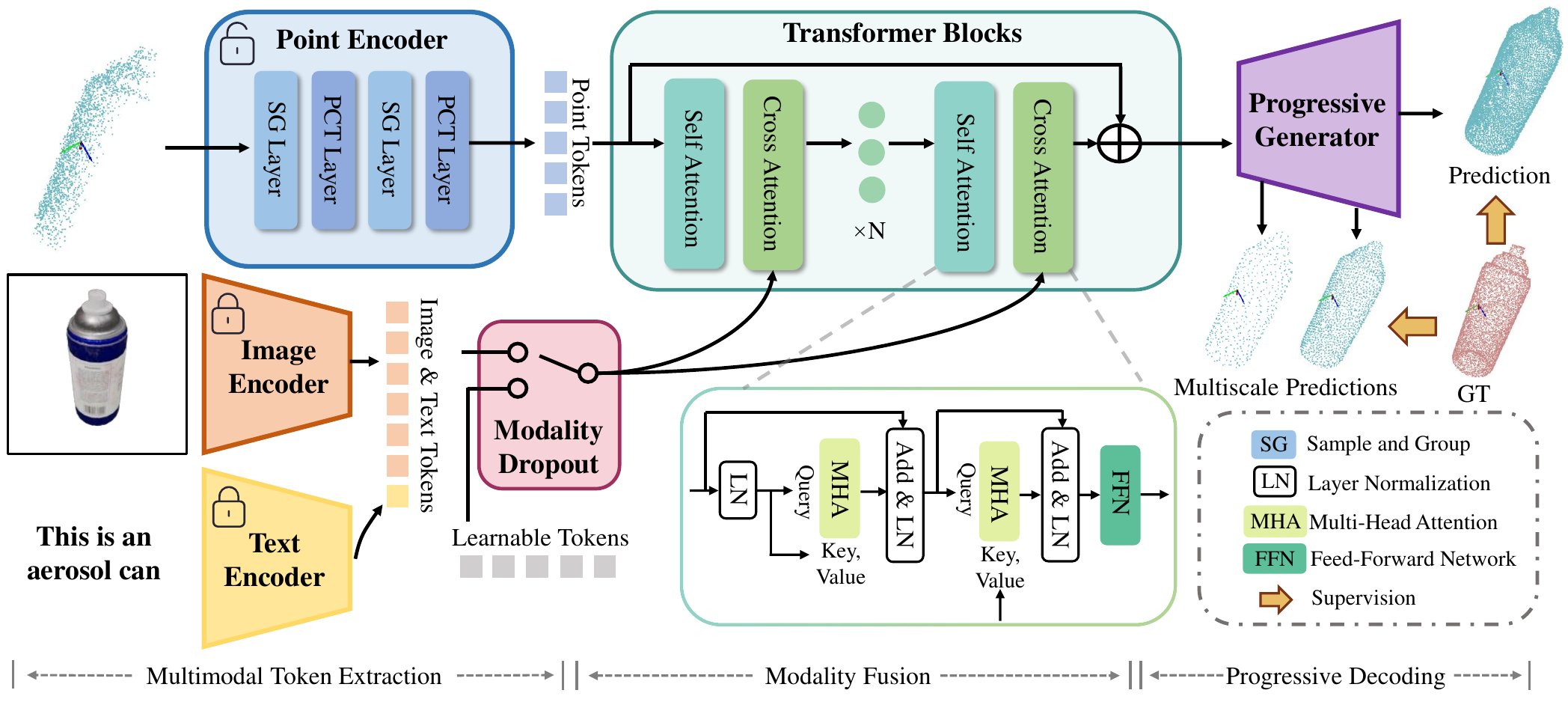}
\caption{Overview of the proposed MGPC architecture. The framework comprises three stages: (i) Multimodal Token Extraction, which encodes point clouds, RGB images, and text into modality-specific tokens; (ii) Modality Fusion, which fuses these tokens using a scalable Transformer with a modality dropout module; and (iii) Progressive Decoding, which generates completed point clouds in a coarse-to-fine manner with multiscale supervision.}
\label{fig2}
\end{figure*}

\section{METHOD}
\subsection{Overview}
Given a partial and sparse point cloud $P \in R^{N_{s} \times 3}$, our goal is to predict the complete point cloud $P_c \in R^{N_{c} \times 3}$ with an RGB image $I \in R^{H \times W \times 3}$ and a text description $T$ provided as auxiliary modalities. Here, $N_{s}$ and $N_{c}$ denote the number of input and output points, respectively, while $H$ and $W$ represent the image height and width.

As illustrated in Fig. \ref{fig2}, the proposed architecture consists of three stages: Multimodal Token Extraction, Modality Fusion, and Progressive Decoding. Tokens from different modalities are first extracted by modality-specific encoders, then refined and fused through a scalable Transformer architecture equipped with a modality dropout module. Finally, a progressive generator predicts multiscale completed point clouds in a coarse-to-fine manner.

\subsection{Multimodal Token Extraction}
We employ modality-specific encoders to extract tokens from the input point cloud, image, and text. Specifically, a lightweight point encoder produces point tokens $T_p \in R^{N_{p} \times d}$, where $d$ denotes the token dimension and $N_p$ is the number of downsampled points. Unlike previous approaches \cite{pcn2018, pointr2021, adapointr2023} that adopt a shared bottleneck architecture, we observe that aggressive dimensionality reduction of point features often leads to the loss of geometric details. In order to preserve richer structural information, we adopt a higher-resolution encoder based on KNN aggregation and self-attention mechanisms \cite{pct2021}. 

For the image and text modalities, we employ DINOv2 \cite{dinov22023} and CLIP \cite{clip2021} as pretrained encoders and keep their weights frozen during training, thereby leveraging large-scale internet knowledge to enhance generalization. These encoders can be readily replaced by other modern feature extractors, such as SigLIP2 \cite{siglip2025}. Image tokens $T_i \in R^{N_{i} \times d}$ and the text token $T_t \in R^{d}$ are obtained by projecting the original features into the shared embedding space via lightweight projection heads, where $N_{i}$ denotes the number of image patches.

\textbf{Modality Dropout.} Dropout \cite{dropout2014} is a widely used regularization technique for improving generalization. Motivated by this principle, we propose a modality dropout strategy to explicitly regularize multimodal conditioning and reduce the reliance on auxiliary inputs. Formally, the condition tokens $T_c$ are defined as:
\begin{equation}
\label{formula1}
T_c = 
\begin{cases}
    [T_i, ~ T_t], & \text{if} ~ z \geq p, \\
    T_{learn}, & \text{if} ~ z < p,
\end{cases}
\end{equation}
where $[\cdot,\cdot]$ denotes concatenation, $z \sim \mathcal{U}(0, 1)$, and $T_{learn}$ is a set of learnable placeholder tokens with matched dimensionality to $[T_i, ~ T_t]$. In practice, $T_{learn}$ is parameterized as trainable embeddings (initialized from a standard normal distribution) and optimized jointly with the network, serving as an empty-condition representation when auxiliary modalities are unavailable. During training, multimodal tokens are randomly replaced by $T_{learn}$, forcing the model to perform completion both with and without auxiliary modalities. This stochastic conditioning acts as an effective regularizer, mitigating overfitting to visual and semantic cues and encouraging the backbone to preserve strong geometric reasoning from the partial point cloud. At inference time, image and text are treated as optional inputs. When modalities are unavailable, $T_{learn}$ is used as the condition tokens, enabling flexible deployment without architectural changes.

\subsection{Modality Fusion}
Many existing point cloud completion methods introduce strong geometric priors or local operators, such as KNN-based aggregation \cite{pointr2021, proxyformer2023, snowflakenet2021, seedformer2022} or handcrafted shape priors \cite{crn2020, odgnet2024}. While effective in limited data regimes, such inductive biases may restrict scalability and generalization as dataset size increases \cite{ptv32024}. Moreover, modality-specific designs are often difficult to extend to other input modalities. Following the recent trend in large-scale 3D reconstruction \cite{mcc2023, lrm2023}, we adopt a general-purpose Transformer architecture with self-attention and cross-attention to fuse multimodal tokens in a unified way. 

Each transformer block consists of a self-attention layer, a cross-attention layer, and a Feed-Forward Network (FFN). The self-attention operation refines point tokens $T_p$ as:
\begin{equation}
\label{formula2}
T^{'}_{p} = \text{MHA}(\text{LN}(T_p), \text{LN}(T_p), \text{LN}(T_p)) + T_p,
\end{equation}
where MHA denotes standard multi-head attention, and LN is the layer normalization. The refined point tokens $T^{'}_{p}$ are then fused with condition tokens through cross attention:
\begin{equation}
\label{formula3}
T^{''}_{p} = \text{MHA}(\text{LN}(T^{'}_{p}), \text{LN}(T_c), \text{LN}(T_c)) + T^{'}_{p}
\end{equation}
Finally, an FFN with residual connections projects the features to the output space. The Transformer block is stacked $N$ times, followed by an additional residual connection across blocks.

\begin{figure*}[!t]
\centering
\includegraphics[width=5in]{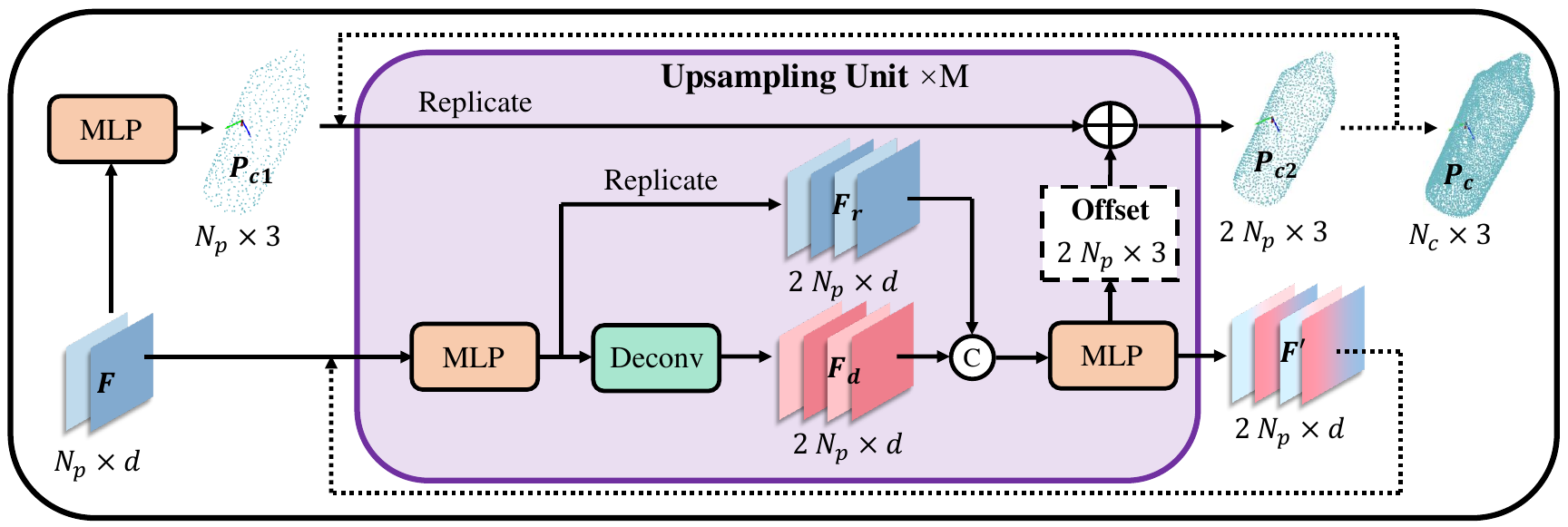}
\caption{Illustration of the proposed progressive generator. Deconvolution and feature replication operations are used to regress point offsets for point cloud upsampling. The process is repeated to generate completed point clouds at multiple scales.}
\label{fig3}
\end{figure*}

\subsection{Progressive Decoding}
The generator maps fused features back to 3D coordinates, which directly impacts the quality of completed point clouds \cite{letup2024}. Previous methods commonly adopt folding-based or MLP-based generators \cite{pcn2018, pointr2021, adapointr2023}, which either deform a 2D grid into 3D surfaces or regress coordinates in a single step. While effective in simple settings, these generators often have limited generative capacity to represent complex and multimodal geometric distributions, and the one-shot regression paradigm can be sensitive to missing observations and prone to producing uneven point densities. Inspired by point cloud upsampling literature \cite{pucrn2022, hfcipu2024}, we design a progressive generator that performs coarse-to-fine reconstruction through iterative refinement.

As shown in Fig. \ref{fig3}, the fused features $F \in R^{N_{p} \times d}$ are first mapped to a coarse prediction $P_{c1} \in R^{N_{p} \times 3}$ via an MLP. We then apply a sequence of upsampling units, each doubling the point density, to progressively enrich local details and increase resolution. In each unit, features are expanded to $2 \, N_{p} \times d$ through two complementary branches: a deconvolution branch that performs learned feature splitting to introduce local variation, and a replication branch that preserves global structure and stabilizes the expansion. Instead of directly regressing absolute coordinates at higher resolution, the generator predicts relative offsets conditioned on the expanded features. This residual formulation constrains updates to local neighborhoods around the coarse geometry, which simplifies optimization, reduces geometric drift, and encourages a more uniform spatial distribution. The predicted offsets are added to the replicated point coordinates to obtain an intermediate prediction $P_{c2} \in R^{2 \, N_{p} \times 3}$. Repeating this process for $M$ units yields the final output $P_{c} \in R^{N_c \times 3}$.

\subsection{Loss Function}
Chamfer Distance (CD) \cite{cd2017} is a commonly used loss function. We adopt HyperCD \cite{hypercd2023}, an improved variant of the CD, which computes distances in hyperbolic space to suppress outliers. The HyperCD loss between prediction $P_c$ and ground truth $G$ is defined as:
\begin{multline}
    \label{formula4}
    \mathcal{L}_{HCD}(P_c, G) = \frac{1}{|P_c|}\sum_{p \in P_c} \text{arcosh}(1+ \alpha \, \min_{g \in G}||p-g||) 
    \\+ \frac{1}{|G|}\sum_{g \in G} \text{arcosh}(1+ \alpha \, \min_{p \in P_c}||g-p||),
\end{multline}
where $\alpha=1$ following \cite{hypercd2023}.

Given multiscale predictions and their corresponding ground truths $\{(P_{ci}, G_i)|i=1, 2, \dots, K \}$, the overall loss is:
\begin{equation}
\label{formula5}
\mathcal{L} = \sum_{i=1}^{K} \beta_{i} \mathcal{L}_{HCD}(P_{ci}, G_i)
\end{equation}
where $\beta_i$ balances losses at different scales.

\begin{figure*}[!t]
\centering
\includegraphics[width=0.95\textwidth]{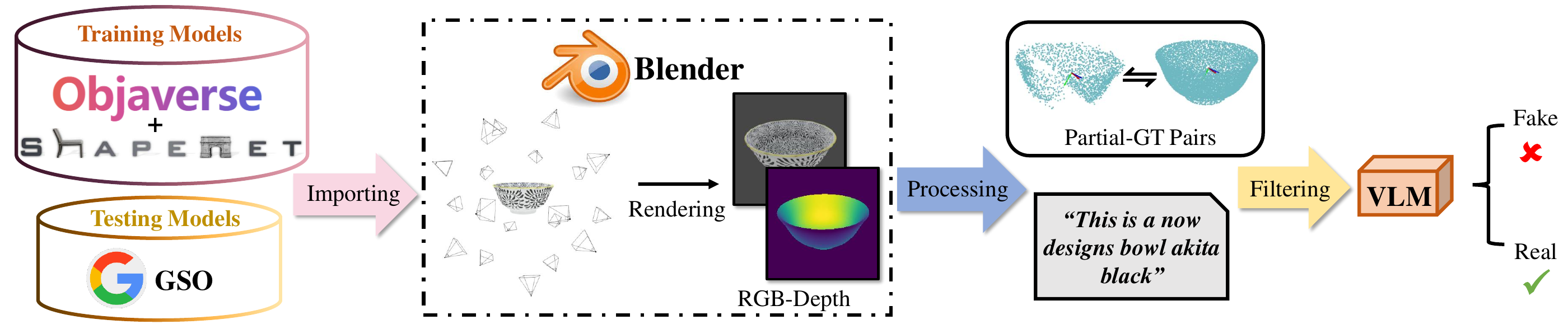}
\caption{Overview of the proposed data generation pipeline. Mesh models from Objaverse, ShapeNet, and GSO are rendered in Blender to produce RGB-D images, which are processed and normalized to form data pairs with text. A VLM-based filter is applied to ensure data quality.}
\label{fig4}
\end{figure*}

\section{DATASET CONSTRUCTION}
All existing datasets for point cloud completion suffer from limited category diversity and data scale \cite{survey22022}. Moreover, many of them adopt settings that deviate from realistic sensing and deployment, such as artificial cropping for partial generation and GT-centric normalization \cite{pdr2021, survey2023}, which substantially hinder the generalization ability of current completion models. To address these limitations, we develop an automated data generation pipeline assisted by a VLM and establish a large-scale benchmark to support comprehensive evaluation and improved generalization. 

\subsection{Generation Pipeline}
As illustrated in Fig. \ref{fig4}, all 3D models are collected from the large-scale Objaverse dataset \cite{objaverse2023}, the real-world GSO dataset \cite{gso2022}, and the classical ShapeNet dataset \cite{shapenet2015}. To ensure high-quality data generation, Blender\footnote{www.blender.org} is employed as the physical engine. RGB images and depth maps are rendered via path tracing from the virtual cameras placed at 20 viewpoints uniformly distributed on a Fibonacci grid \cite{mathographics1991}, enabling comprehensive coverage of object surfaces. Unlike previous approaches that generate partial point clouds through naive cropping or occlusion operations \cite{pointr2021, vipc2021}, we simulate realistic sensing processes by back-projecting depth maps and injecting noise to mimic real-world depth cameras and LiDAR sensors. The complete ground-truth point cloud is uniformly sampled using the FPS algorithm. In addition, text descriptions are automatically extracted from object tags or category labels to support multimodal learning settings.

\textbf{Normalization.} Most existing datasets adopt GT-centric normalization for simplicity \cite{pcn2018, pointr2021, vipc2021}, where both partial and complete point clouds are normalized using the centroid and scale of the ground-truth shape. However, such a protocol is impractical in real-world scenarios, as ground-truth geometry is unavailable during inference. To resolve this inconsistency, we normalize each data pair based on the centroid and scale of the input partial point cloud instead. These normalization parameters are stored as part of the dataset, ensuring consistency between training and real-world deployment. 

\textbf{Filtering.} Despite its large scale and diversity, Objaverse contains a non-negligible portion of low-quality 3D models due to the lack of manual verification \cite{openshape2023}. Therefore, additional filtering is essential. Leveraging the strong reasoning and commonsense capabilities of modern vision–language models such as GPT-4V\footnote{www.openai.com}, we design a VLM-based data selector to automatically filter out low-quality samples. Specifically, a predefined prompt template is populated with rendered images and semantic tags of each shape, and the VLM is asked to judge its validity. Through this process, approximately 5\% of models with unrealistic geometries, corrupted formats, or abnormal structures are removed. We further perform manual inspection on randomly sampled subsets to ensure overall data quality. The entire data generation pipeline for our MGPC-1M takes approximately two weeks using four GPUs.

\subsection{MGPC-1M Benchmark}
As summarized in Table \ref{tab1}, the training split of MGPC-1M covers over 1k categories spanning a broad range of daily objects, offering substantially greater category diversity than existing point cloud completion datasets. MGPC-1M contains over one million training pairs, exceeding the combined scale of prior benchmarks in this domain. Unlike most existing datasets that provide only point clouds, MGPC-1M additionally includes RGB images and text descriptions, supporting multimodal evaluation and learning. Beyond scale, MGPC-1M adopts more realistic data generation protocols, including depth back-projection with sensor-like noise and input-centric normalization based on the partial observation, which better matches real-world deployment where ground-truth geometry is unavailable. In contrast to datasets that rely solely on synthetic data, our test split is constructed from the real-world GSO dataset and is entirely disjoint from the training data, enabling rigorous evaluation of cross-domain generalization \cite{lrm2023}. The resulting test set contains 1,031 novel instances with high-quality real-world models.

\begin{table*}[!t]
\centering
\caption{Comparison with Single-Modal Baselines on Our Dataset in Terms of Chamfer Distances $\times$ 1000 (Lower is Better) and F-Score@1\% (Higher is Better)}
\label{tab2}
\setlength{\tabcolsep}{4.5pt}
\begin{tabular}{c|ccc|ccc|ccc|ccc|ccc}
\hline
\multirow{2}{*}{Methods} & \multicolumn{3}{c|}{Average}                         & \multicolumn{3}{c|}{Shoes}                        & \multicolumn{3}{c|}{Action Figures}                 & \multicolumn{3}{c|}{Bags}                          & \multicolumn{3}{c}{Electronic Devices}           \\ \cline{2-16} 
                         & CD-$\ell_2$          & CD-$\ell_1$          & F1             & CD-$\ell_2$          & CD-$\ell_1$          & F1             & CD-$\ell_2$          & CD-$\ell_1$          & F1             & CD-$\ell_2$          & CD-$\ell_1$           & F1             & CD-$\ell_2$          & CD-$\ell_1$          & F1             \\ \hline
PCN \cite{pcn2018}    & 0.766          & 12.38         & 0.567          & 0.811          & 13.29         & 0.518          & 0.777          & 13.90         & 0.491          & 1.049          & 13.68          & 0.565          & 0.785          & 11.27         & 0.637          \\ \cline{1-1}
PoinTr \cite{pointr2021}    & 0.485          & 10.11         & 0.681          & 0.585          & 11.43         & 0.587          & 0.551          & 10.46         & 0.657          & 1.005          & 12.82          & 0.613          & 0.471          & 9.12          & 0.730          \\ \cline{1-1}
SnowFlakeNet \cite{snowflakenet2021}  & 0.510          & 10.50         & 0.651          & 0.629          & 12.00         & 0.564          & 0.491          & 10.81         & 0.629          & 0.869          & 12.97          & 0.583          & 0.489          & 9.57          & 0.704          \\ \cline{1-1}
SeedFormer \cite{seedformer2022}               & 0.463          & 9.85          & 0.678          & 0.592          & 11.38         & 0.597          & 0.419          & 9.70          & 0.695          & 0.783          & 12.21          & 0.616          & 0.429          & 8.90          & 0.747          \\ \cline{1-1}
PointAttn \cite{pointattn2024}                & 0.453          & 9.66          & 0.704          & 0.627          & 11.37         & 0.593          & 0.477          & 10.26         & 0.663          & 0.806          & 12.16          & 0.628          & 0.407          & 8.56          & 0.762          \\ \cline{1-1}
AdaPoinTr \cite{adapointr2023}                & 0.408          & 9.03          & 0.751          & 0.530          & 10.43         & 0.649          & 0.377          & 8.96          & 0.746          & 0.887          & 11.52          & 0.675          & 0.415          & 8.06          & 0.798          \\ \hline
Ours                     & \textbf{0.378} & \textbf{8.57} & \textbf{0.772} & \textbf{0.411} & \textbf{9.95} & \textbf{0.679} & \textbf{0.344} & \textbf{8.694} & \textbf{0.764} & \textbf{0.605} & \textbf{10.68} & \textbf{0.688} & \textbf{0.285} & \textbf{7.43} & \textbf{0.831} \\ \hline
\multicolumn{16}{l}{\footnotesize
\textit{Note:}
Lower is better for CD; higher is better for F-Score. The best is highlighted in bold.
}
\end{tabular}
\end{table*}

\begin{table*}[!t]
\centering
\caption{Comparison with Cross-Modal Baselines on Our Dataset in Terms of Chamfer Distances $\times$ 1000 (Lower is Better) and F-Score@1\% (Higher is Better)}
\label{tab3}
\setlength{\tabcolsep}{4.5pt}
\begin{tabular}{c|ccc|ccc|ccc|ccc|ccc}
\hline
\multirow{2}{*}{Methods} & \multicolumn{3}{c|}{Average}                         & \multicolumn{3}{c|}{Shoes}                        & \multicolumn{3}{c|}{Action Figures}              & \multicolumn{3}{c|}{Bags}                         & \multicolumn{3}{c}{Electronic Devices}           \\ \cline{2-16} 
                         & CD-$\ell_2$          & CD-$\ell_1$          & F1             & CD-$\ell_2$          & CD-$\ell_1$          & F1             & CD-$\ell_2$          & CD-$\ell_1$          & F1             & CD-$\ell_2$          & CD-$\ell_1$          & F1             & CD-$\ell_2$          & CD-$\ell_1$          & F1             \\ \hline
CSDN \cite{csdn2023}                     & 1.880          & 21.29          & 0.233          & 2.277          & 23.14          & 0.197          & 1.721          & 20.94          & 0.272          & 3.235          & 25.42          & 0.184          & 1.870          & 19.43          & 0.288          \\ \cline{1-1}
EGIInet \cite{egiinet2024}                  & 1.018          & 16.90          & 0.289          & 1.305          & 18.27          & 0.252          & 0.945          & 16.29          & 0.328          & 1.580          & 19.61          & 0.243          & 1.072          & 15.56          & 0.355          \\ \cline{1-1}
XMFNet \cite{xmfnet2022}                  & 0.922          & 16.43          & 0.284          & 1.111          & 18.03          & 0.241          & 0.881          & 16.18          & 0.315          & 1.296          & 18.70          & 0.232          & 0.725          & 14.64          & 0.350          \\ \hline
Ours                     & \textbf{0.623} & \textbf{12.53} & \textbf{0.495} & \textbf{0.732} & \textbf{14.28} & \textbf{0.408} & \textbf{0.564} & \textbf{12.30} & \textbf{0.528} & \textbf{1.174} & \textbf{14.95} & \textbf{0.417} & \textbf{0.551} & \textbf{11.09} & \textbf{0.587} \\ \hline
\end{tabular}
\end{table*}

\begin{figure*}[!t]
\centering
\includegraphics[width=7in]{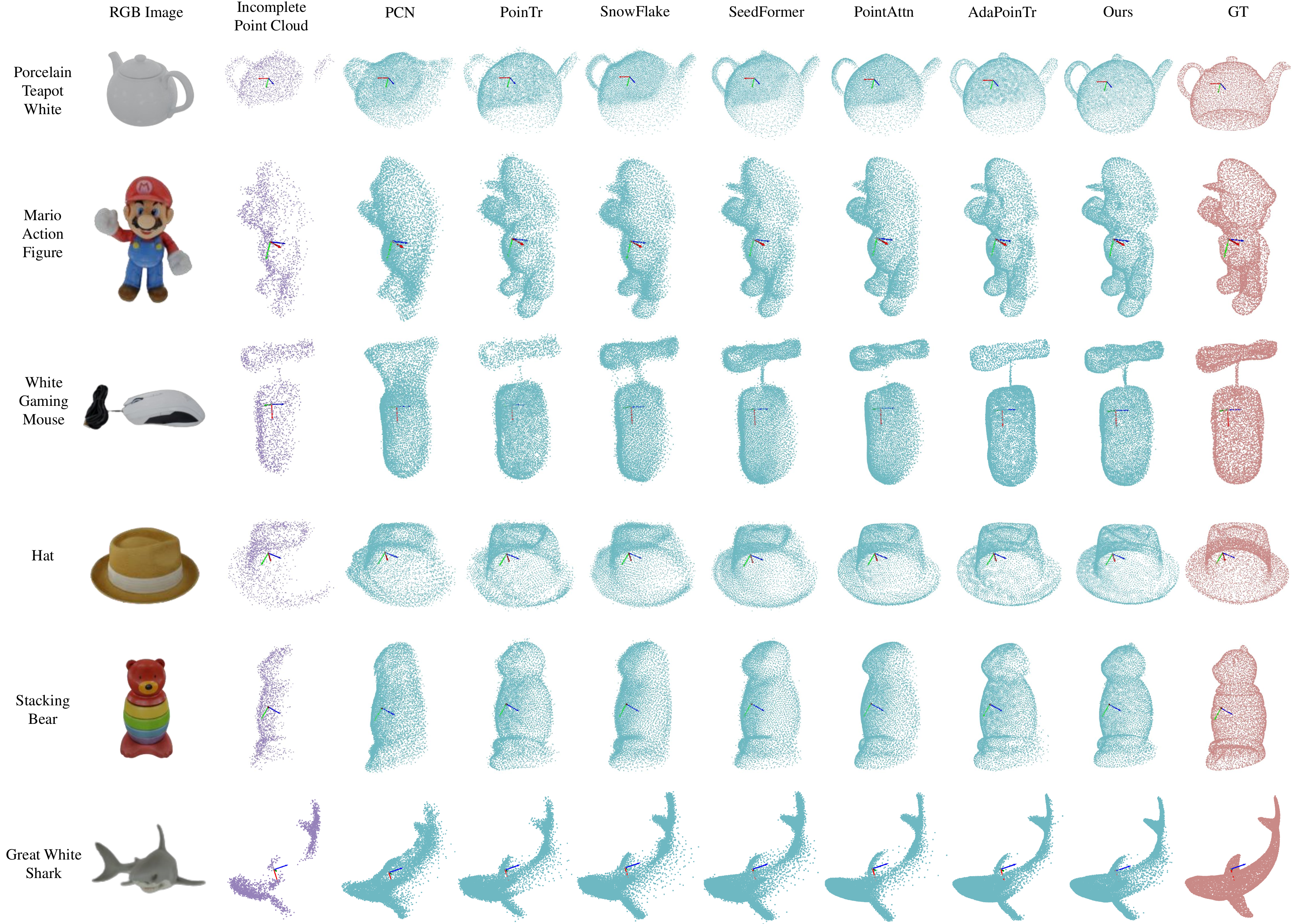}
\caption{Visual comparison with single-modal baselines using 8192 output points on MGPC-1M. (Best viewed magnified.)}
\label{fig5}
\end{figure*}

\begin{figure*}[!t]
\centering
\includegraphics[width=6in]{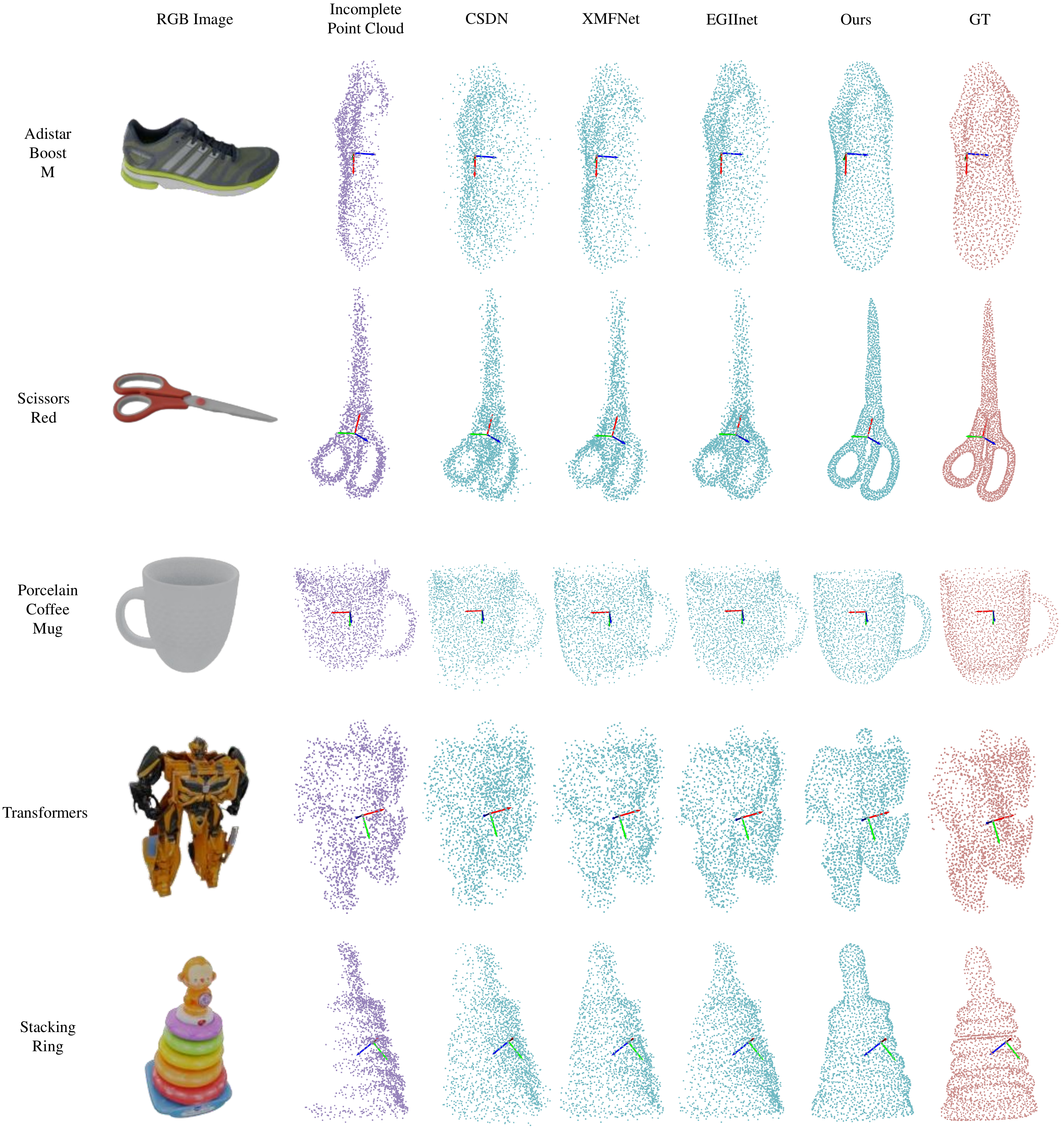}
\caption{Visual comparison with cross-modal baselines using 2048 output points on MGPC-1M.}
\label{fig6}
\end{figure*}

\section{EXPERIMENTS}
\subsection{Evaluation Metrics}
We follow standard protocols in prior works \cite{pointr2021, seedformer2022, adapointr2023} and employ Chamfer Distances \cite{cd2017} and F-Score@1\% \cite{pointr2021}. Depending on the distance norm, CD is reported as CD-$\ell_1$ and CD-$\ell_2$. While CD-$\ell_1$ and CD-$\ell_2$ measure the overall geometric discrepancy between the reconstructed point clouds and the ground truth, the F-Score evaluates surface-level similarity between two point sets. All methods are evaluated on the full test set using the same batch size. In addition, we report category-level results on four manually selected categories, namely shoes, action figures, bags, and electronic devices, containing approximately 5k, 1k, 600, and 679 samples, respectively.

\subsection{Implementation Details}
\subsubsection{Hyper-parameters}
Following prior works, the number of input points is fixed to 2048. To ensure fair comparisons with both single-modal methods \cite{pcn2018, pointr2021} and cross-modal approaches \cite{xmfnet2022}, our network is trained to produce 8192 and 2048 output points, respectively. The modalities fusion module consists of 8 transformer blocks, and the progressive decoder contains 2 upsampling units. To match common settings in modern vision backbones, the rendered RGB images have a resolution of 224 $\times$ 224.

\subsubsection{Training Details}
Our model is implemented in PyTorch and optimized using AdamW with an initial learning rate of $1 \times 10^{-4}$. We train our model end-to-end for 100 epochs on 4 RTX 4090 GPUs with a batch size of 128. For consistency and fairness, all methods are trained on the full training set of MGPC-1M containing over 1,000 categories and more than one million data pairs.

\subsubsection{Baselines} 
We select several representative and recent state-of-the-art methods as baselines. Single-modal approaches include PCN \cite{pcn2018}, PoinTr \cite{pointr2021}, SnowFlakeNet \cite{snowflakenet2021}, SeedFormer \cite{seedformer2022}, PointAttn \cite{pointattn2024}, and AdaPoinTr \cite{adapointr2023}, all evaluated with 8192 output points. We further compare with cross-modal completion methods, including XMFNet \cite{xmfnet2022}, CSDN \cite{csdn2023} and EGIInet \cite{egiinet2024}, which generate 2048-point outputs. All baselines are reimplemented within our MGPC-1M using their official open-source code and original settings to ensure fair comparisons.

\begin{figure*}[!t]
\centering
\includegraphics[width=0.95\textwidth]{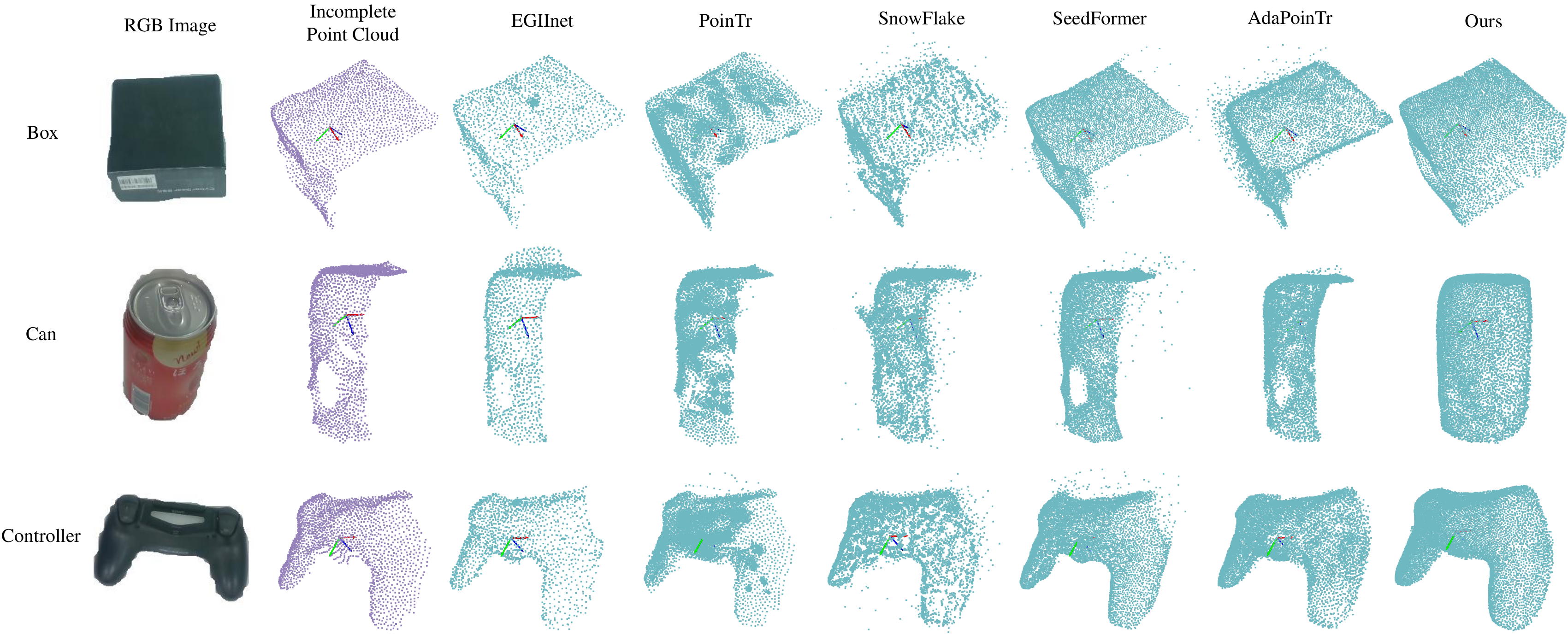}
\caption{Qualitative results demonstrating zero-shot generalization on in-the-wild real-world data.}
\label{fig7}
\end{figure*}

\subsection{Evaluation on MGPC-1M}

To validate the effectiveness of the proposed method, we conduct extensive quantitative and qualitative evaluations on our MGPC-1M.

\subsubsection{Numerical Results} 
Table \ref{tab2} reports quantitative results with 8192 output points. Our method consistently outperforms all single-modal baselines by a substantial margin across all metrics, establishing a new state of the art. Table \ref{tab3} further compares our approach with cross-modal methods, where our model also achieves superior performance on all evaluation metrics. In particular, compared with the second-best method, XMFNet, our approach achieves relative improvements of approximately 32\%, 23\%, and 74\% in CD-$\ell_2$, CD-$\ell_1$, and F-Score, respectively.

\subsubsection{Visual Results} 
Fig. \ref{fig5} presents qualitative comparisons between our method and representative single-modal baselines. All point clouds, except for the inputs, contain 8192 points and are manually rotated for clearer visualization. The red, green, and blue axes indicate the x, y, and z axes, respectively, with the camera oriented along the z-axis. As shown, our multimodal approach produces more accurate and visually coherent reconstructions than previous methods. The previous point-based approach, PCN, fails to recover complete structures from partial inputs. Transformer-based methods, including PoinTr, SnowFlake, and Seedformer tend to generate noise and outliers, likely due to strong geometric inductive biases. While PointAttn and AdaPoinTr yield improved results, they still suffer from loss of fine-grained details due to limitations in their generation modules. In contrast, benefiting from multimodal information and a scalable architecture, our method generates more complete and sharper reconstructions. Fine yet critical details, such as the teapot spout in the first row, Mario’s hand in the second row, and the mouse cable in the third row, are accurately recovered. Furthermore, owing to the proposed progressive decoder, the generated point clouds exhibit a more uniform spatial distribution, whereas previous methods are often biased toward the observed partial regions.

Fig. \ref{fig6} further shows qualitative comparisons with cross-modal methods using 2048-point outputs on novel objects. Our approach consistently produces completions that better preserve global shape and local geometric structures across both simple objects (e.g., scissors and mugs) and complex shapes (e.g., Transformers). Notably, despite the dedicated RGB image fusion designs adopted by CSDN, XMFNet, and EGIINet, their results can still be sensitive to the modality gap between 2D appearance and 3D geometry, which may manifest as incomplete recovery in occluded regions. By contrast, our unified attention-based fusion allows point tokens to remain the primary geometric carrier while selectively incorporating auxiliary visual and semantic cues, thereby reducing ambiguity and generating more coherent completions.

\subsection{Zero-Shot Evaluation on In-the-Wild Data}

While existing approaches achieve promising performance on curated benchmarks, their effectiveness on in-the-wild data remains largely unexplored. To evaluate the real-world generalization ability and practical applicability of our method, we conduct, to the best of our knowledge, the first in-the-wild evaluation for point cloud completion.
 
RGB images and depth maps are captured using an Intel RealSense D435i camera. We employ the open-vocabulary detection and segmentation method GroundingDINO \cite{groundingdino2024} to localize target objects and obtain corresponding RGB and depth segments. The segmented depth maps are back-projected into point clouds, followed by Statistical Outlier Removal (SOR) algorithm to eliminate background noise and spurious points. The resulting point clouds are then normalized to zero centroid and unit scale to form incomplete inputs.

Fig. \ref{fig7} presents qualitative comparisons with both single-modal baselines \cite{pointr2021, snowflakenet2021, seedformer2022, adapointr2023} and a cross-modal \cite{egiinet2024} baseline, all evaluated using their official pretrained models. Note that RGB images are used by the EGIInet \cite{egiinet2024} and our model. The partial inputs and predictions of EGIInet contain 2048 points, while other point clouds consist of 8192 points. As shown, existing methods generally fail to recover complete object geometries in unseen real-world scenarios, leaving large missing regions in areas without observations. We attribute these limitations to several factors, including architectural constraints, insufficient training data diversity, and inconsistent normalization strategies. Even for geometrically simple objects such as boxes, baseline methods tend to produce excessive noise due to model uncertainty. For more complex objects, such as the controller, they struggle to reconstruct fine-grained structures, including corners. Moreover, methods such as SnowFlakeNet and PoinTr often generate clustered and non-uniform point distributions, which can be attributed to strong geometric priors and limited decoder capacity. In contrast, our method produces more complete and structurally consistent reconstructions for novel real-world objects. Benefiting from a large-scale, diverse dataset and a scalable multimodal architecture, our predictions exhibit superior quality in terms of completeness, detail preservation, and spatial uniformity. In addition, our approach demonstrates robustness to sensor artifacts. For instance, the point cloud of the can contains a hole near the front surface due to inherent limitations of depth sensing. With the aid of multimodal auxiliary information, our model successfully infers and repairs this missing region, whereas other methods fail to address the defect.

\begin{table}[!t]
\centering
\caption{Ablation Study on Different Multimodal Input Configurations}
\label{tab4}
\begin{tabular}{l|ccc}
\hline
Variations            & CD-$\ell_2$ ($\downarrow$) & CD-$\ell_1$ ($\downarrow$) & F1 ($\uparrow$) \\ \hline
w/o Image \& Text & 0.458      & 9.72      & 0.722             \\ \hline
w/o Image            & 0.423      & 9.21      & 0.745             \\ \hline
w/o Text             & 0.395      & 9.02      & 0.752             \\ \hline
Ours                  & \textbf{0.378}      & \textbf{8.57}      & \textbf{0.772}             \\ \hline
\end{tabular}
\end{table}


\begin{table}[!t]
\centering
\caption{Ablation Study on the Modality Dropout Module}
\label{tab5}
\begin{tabular}{cc|ccc}
\hline
\multicolumn{2}{c|}{Variations}                          & CD-$\ell_2$ ($\downarrow$)            & CD-$\ell_1$ ($\downarrow$)           & F1 ($\uparrow$)            \\ \hline
\multicolumn{1}{c|}{\multirow{5}{*}{Probability $p$}} & 0.00 & 0.399          & 8.85          & 0.757          \\ \cline{2-5} 
\multicolumn{1}{c|}{}                             & 0.25 & 0.385          & 8.68          & 0.764          \\ \cline{2-5} 
\multicolumn{1}{c|}{}                             & 0.50 & \textbf{0.378} & \textbf{8.57} & \textbf{0.772} \\ \cline{2-5} 
\multicolumn{1}{c|}{}                             & 0.75 & 0.431          & 9.29          & 0.739          \\ \cline{2-5} 
\multicolumn{1}{c|}{}                             & 1.00 & 0.458          & 9.72          & 0.722          \\ \hline
\multicolumn{2}{c|}{Modalities Missing}                  &  0.470              & 9.10              & 0.749               \\ \hline
\end{tabular}
\end{table}

\begin{table}[!t]
\centering
\caption{Ablation Study on Decoder and Transformer Design Choices}
\label{tab6}
\begin{tabular}{l|ccc}
\hline
Variations               & CD-$\ell_2$ ($\downarrow$)         & CD-$\ell_1$ ($\downarrow$)         & F1 ($\uparrow$)         \\ \hline
w/ Folding-Based Decoder  & 0.391                & 8.82                     & 0.751                \\ \hline
w/ MLP-Based Decoder      & 0.385                & 8.79                     & 0.748                     \\ \hline
w/ Self-Attention Layers  & 0.382   & 8.65   & 0.767 \\ \hline
Ours                     & \textbf{0.378}       & \textbf{8.57}       & \textbf{0.772}       \\ \hline
\end{tabular}
\end{table}

\begin{figure}[!t]
\centering
\includegraphics[]{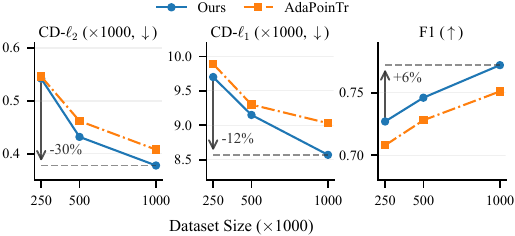}
\caption{Scaling analysis. The horizontal axis denotes the number of training pairs (in thousands), and the vertical axis reports CD-$\ell_2$, CD-$\ell_1$ (×1000, lower is better), and F-Score (higher is better) for Ours and AdaPoinTr.}
\label{fig8}
\end{figure}

\subsection{Ablation Studies}

We conduct a series of ablation studies to analyze the contribution of key components in our framework. All experiments are performed on the proposed large-scale dataset with 8192 output points and the same batch size.

\subsubsection{Multimodal Inputs} 
We first examine the impact of different input modalities, as summarized in Table \ref{tab4}. In the first setting, both image and text encoders are removed, and the multimodal tokens in the Transformer blocks are replaced with learnable tokens, resulting in a purely single-modal network. As expected, all evaluation metrics degrade significantly, indicating the importance of semantic and visual information provided by additional modalities. In the second and third settings, we remove either the image or text modality. The performance degradation in these cases is notably smaller than that of the single-modal variant, demonstrating that incorporating more modalities leads to stronger performance gains. Furthermore, the configuration using RGB images yields slightly better results than that using text alone, suggesting that images provide richer appearance cues.

\subsubsection{Module Designs} 
To validate the effectiveness of the proposed modality dropout module, we vary the dropout probability $p$ from 0.0 to 1.0 and retrain the model, as reported in Table \ref{tab5}. When $p=0.0$, the module is effectively disabled. As $p$ increases to 0.5, overall performance consistently improves, indicating that the introduced stochasticity alleviates overfitting and enhances generalization. However, when $p$ exceeds 0.5, performance begins to decline, likely because excessive reliance on learnable tokens weakens the contribution of multimodal information. Based on these observations, we set $p=0.5$ in all experiments. In addition, we evaluate robustness under missing-modality settings enabled by modality dropout. Specifically, we remove both image and text inputs at inference time and report the resulting performance (last row of Table \ref{tab5}). Although this setting incurs a moderate performance drop relative to the full multimodal configuration, it remains competitive, demonstrating that the proposed mechanism supports flexible inference under varying modality availability.

We further evaluate the design of the decoder in Table \ref{tab6}. In the first two rows, we replace the proposed progressive generator with a folding-based decoder \cite{pointr2021} and MLP-based decoder \cite{adapointr2023}, respectively. Both alternatives lead to notable performance degradation, reflecting their limited capacity to model complex point cloud distributions. In contrast, our progressive decoder, which combines coarse-to-fine generation with deconvolution and replication operations, achieves substantially better results. In the third ablation, we replace the cross-attention layers in the modality fusion module with self-attention layers and concatenate point and multimodal tokens directly. This modification results in inferior performance, highlighting the importance of explicitly separating queries, keys, and values and allowing point tokens to dominate the fusion process.

\subsection{Scaling Analysis} 
We investigate the impact of training data scale in Fig. \ref{fig8}. We randomly sample 250k and 500k data pairs from the MGPC-1M dataset and train both our model and the second-best baseline, AdaPoinTr, on these subsets. As the dataset size increases, all evaluation metrics improve steadily. Specifically, CD-$\ell_2$ and CD-$\ell_1$ decrease by approximately 30\% and 12\%, respectively, while the F-Score improves by 6\%, demonstrating the effectiveness of our large-scale dataset. Moreover, our method benefits more significantly from data scaling than the baseline, indicating superior scalability. The observed trends also suggest that further expansion of the dataset is likely to yield additional performance gains.

\section{Conclusion}
In this paper, we present a multimodal framework for generalizable point cloud completion that jointly leverages geometric and semantic cues from point clouds, RGB images, and text within a unified Transformer-based architecture. A modality dropout strategy is introduced to improve robustness under varying modality availability, and a progressive decoder is designed to enable coarse-to-fine reconstruction with improved structural consistency and visual fidelity. To facilitate scalable learning and evaluation, we further develop a VLM-assisted data generation pipeline and construct a large-scale benchmark covering over 1,000 categories and one million training pairs. Extensive experiments on both the benchmark and in-the-wild real-world data demonstrate that the proposed approach consistently improves reconstruction quality and exhibits strong zero-shot generalization, highlighting its effectiveness in enhancing the accuracy and robustness of 3D measurement and reconstruction under real-world sensing conditions.



\bibliographystyle{IEEEtran}
\bibliography{ref}


\vspace{11pt}




\vfill

\end{document}